\documentclass[USenglish,oneside,twocolumn]{article}

\usepackage[utf8]{inputenc}
\usepackage[big]{dgruyter_NEW}
\usepackage{filecontents}
\usepackage{dirtytalk}
\usepackage{booktabs}
\usepackage{mathtools}
\usepackage{amsfonts}
\usepackage{amssymb}
\usepackage{amsthm}
\usepackage{microtype}
\usepackage{subfigure}

\DOI{foobar}

\cclogo{\includegraphics{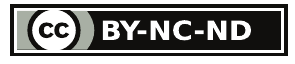}}
  
\begin{document}

  \author[1]{Dmitrii Usynin}

  \author[2]{Daniel Rueckert}

  \author*[3]{Georgios Kaissis}


  \affil[1]{Institute for Artificial Intelligence in Medicine; Department of Diagnostic and Interventional Radiology, Technical University of Munich; Department of Computing, Imperial College London, E-mail: du216@ic.ac.uk}

  \affil[2]{Institute for Artificial Intelligence in Medicine, Technical University of Munich; Department of Computing, Imperial College London; E-mail: d.rueckert@imperial.ac.uk}

  \affil[3]{Department of Diagnostic and Interventional Radiology; Institute for Artificial Intelligence in Medicine, Technical University of Munich; Department of Computing, Imperial College London, Germany E-mail: g.kaissis@tum.de}


  \title{\huge Beyond Gradients: Exploiting Adversarial Priors in Model Inversion Attacks}

  \runningtitle{Beyond Gradients: Exploiting Adversarial Priors in Model Inversion Attacks}


\begin{abstract}
{Collaborative machine learning settings like federated learning can be susceptible to adversarial interference and attacks. One class of such attacks is termed \textit{model inversion attacks}, characterised by the adversary reverse-engineering the model to extract representations and thus disclose the training data. Prior implementations of this attack typically only rely on the captured data (i.e. the shared gradients) and do not exploit the data the adversary themselves control as part of the training consortium. In this work, we propose a novel model inversion framework that builds on the foundations of gradient-based model inversion attacks, but additionally relies on matching the features and the style of the reconstructed image to data that is controlled by an adversary. Our technique outperforms existing gradient-based approaches both qualitatively and quantitatively, while still maintaining the same honest-but-curious threat model, allowing the adversary to obtain enhanced reconstructions while remaining concealed.}
\end{abstract}


  \journalname{Proceedings on Privacy Enhancing Technologies}
\DOI{Editor to enter DOI}
  \startpage{1}
  \received{..}
  \revised{..}
  \accepted{..}

  \journalyear{..}
  \journalvolume{..}
  \journalissue{..}

\maketitle

\section{Introduction}
Machine learning (ML) models have been deployed in a large number of contexts ranging from medical image analysis \cite{chen2019self} to stock market prediction \cite{patel2015predicting}. However, the effective training of such models depends upon the availability of descriptive and representative training data. One paradigm that permits models to leverage larger and more diverse datasets is collaborative machine learning (CML) \cite{verbraeken2020survey}, which permits model training on geographically distributed datasets. CML includes a number of approaches ranging from direct data sharing to transfer learning on publicly available data, all of which allow institutions to share their data with other contributors and thus to train models which generalise better. However, as such data is often sensitive in nature, procuring these datasets directly can be problematic due to data protection and governance regulations which specifically forbid collaborators from exchanging data with each-other. This can be particularly problematic in ML contexts that rely on data that is difficult to obtain (e.g. medical image analysis). Thus, newer paradigms of CML such as federated learning (FL) \cite{konevcny2016federated} have been introduced and enable distributed model training without exchanging the data itself. Instead, ML models are trained locally and only the model \textit{updates} are shared with the rest of the federation of data owners. This approach, however, was shown to be exploitable by adversaries \cite{xie2019dba,melis2019exploiting,nasr2019comprehensive,usynin2021adversarial}, particularly those that can obtain access to shared model updates, as these contain information about the data used to train the model. Such information can be reverse-engineered, allowing the adversary to recover the original data behind the captured update, thus disclosing sensitive information. 

\begin{figure}[h!]
\centering
      \includegraphics[width=0.4\linewidth]{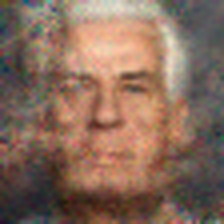}
    \hspace{5mm}%
      \includegraphics[width=0.4\linewidth]{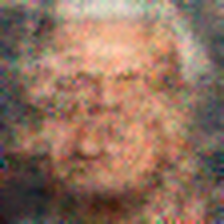}
\caption{Comparison of resulting reconstructions: Our method (left) results in substantially improved reconstruction fidelity compared to the baseline method (right).}
\label{fig:faces_comparison}
\end{figure}

\begin{figure*}[h!]
\centering
\includegraphics[width=0.9\linewidth]{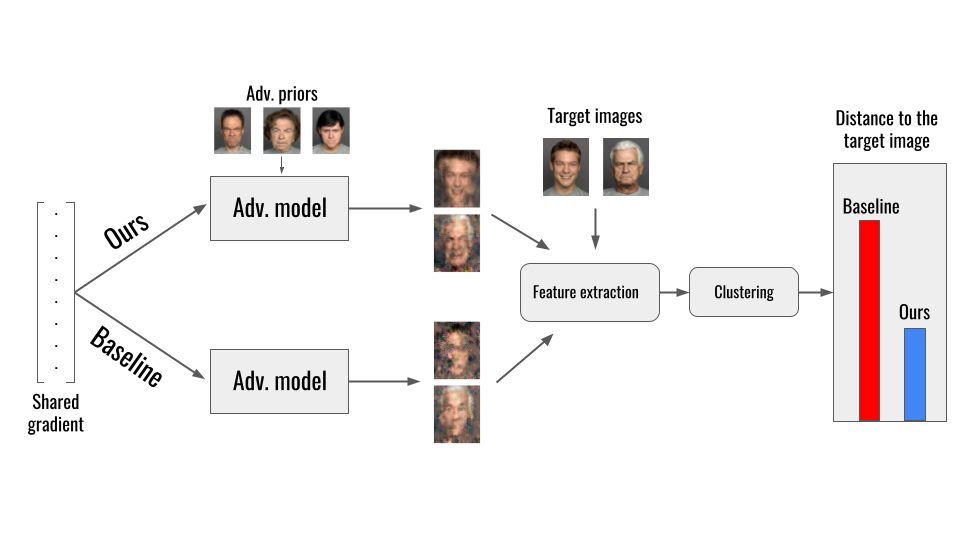}
\caption{Overview of a facial recognition downstream task. The adversarial priors used in our method are sourced from the same distribution as the target data.}
\label{fig:faces_overview}
\end{figure*}

The most prominent class of such attack is termed reconstruction attacks (or model inversion attacks) \cite{fredrikson2015model} which can exploit intermediate model updates in forms of shared activations or gradients. The former was previously presented in He et al. \cite{he2019model}, where authors showed that collaborative inference is vulnerable to an honest-but-curious (HbC) adversary in the form of central server who can invert the shared activations to obtain images on which inference was performed. In this setting, the adversary is assumed to be one of the two collaborating parties and the proposed attack is typically constrained to shallow architectures. Another attack that employs the same threat model was presented by Zhu et al. and termed \textit{deep leakage from gradients} (DLG) \cite{zhu2020deep}. DLG assumes that adversary is HbC and is able to obtain a shared gradient either as the central server or as an adversarial client (when the number of data owners is relatively small). Gradient-based reconstruction attacks can, in comparison to activation-based attacks, obtain images that are either identical or indistinguishable to the ones used to train the model, rendering them a significant threat to a large number of CML implementations. 

While the aforementioned attacks can be threatening to CML implementations such as FL, they are often very brittle in practise and rely on a number of assumptions about the training protocol. Typical assumptions include that the adversary is limited to smaller models, datasets of lower dimensions (e.g. smaller images) or that specific layers are present in the target model (such as normalisation layers). Furthermore, the adversaries in these settings are highly encouraged to perform the attack at the start of the training procedure, while the shared gradient norms are high and, thus, contain more information. These attacks, in addition, both assume that the adversary is part of the training consortium and/or is the aggregation server, which in most cases implies that they have access to or prior knowledge of the training data. However, in practise, gradient-based reconstruction attacks often do not consider this in their choice of the threat model: the adversary does not rely on their own data (that in a number cases can be similar to the data that they are trying to reconstruct). As a result, in a number of contexts, these attacks fail to produce images with high fidelity or only provide incomplete reconstructions. We contend that this is an omission, as the possession of priors with features similar to the training dataset is inherent to many forms of federated learning, and thus their exploitation should be considered. 

In this work, we explore how principled adversarial prior exploitation allows to produce more accurate reconstructions both qualitatively and quantitatively. We analyse a number of learning contexts and model architectures, investigating not only the effects of using such data, but also how the content of the adversarial dataset affects the reconstruction. This issue is significantly more complex in heterogeneous datasets such as the ones used in facial recognition tasks (e.g. FACES \cite{ebner2010faces}) with large intra-class variation \cite{dubey2018pairwise}, making reconstruction a challenging task for the adversary as even if they have the data of the same class, it might not be related to the data of their victim. Alternatively, datasets that contain images where inter-class variation is much smaller (such as abdominal computed tomography (CT) scans in medical image analysis) can be significantly more vulnerable to adversaries that can utilise their own data to facilitate the reconstruction. 

To exploit prior information, we utilise intermediate model activations from attacker-controlled datapoints, which we integrate in form of an additional reconstruction term in gradient-based model inversion setting. Our threat model corresponds to the one of Geiping et al. \cite{geiping2020inverting} (HbC) and the only additional step we take is matching the activations of the image that is being reconstructed to those of the same class controlled by the attacker, similarly to a style transfer loss term from \cite{johnson2016perceptual}. This allows us to maintain the same HbC attacker setting and enhance the reconstructions for which the adversary controls images of the same class as the victim. We show that even an approach where such activations are used as a single, untuned penalty term, outperforms gradient-only reconstruction. We further study the properties of different datasets and architectures when it comes to assessing the severity of our method in various collaborative settings. 

Finally, we discover that our method not only outperforms the existing baseline both quantitatively and qualitatively, but also leads to additional information leakage through a number of adversarial downstream tasks. One such task is facial recognition, where out method is able to produce facial reconstructions that are more similar to the original training data than the existing baseline, resulting in an adversary being able to use such images in \textit{facial spoofing} attacks \cite{ming2020survey, erdogmus2013spoofing} with higher fidelity. An overview of such attack can be found in Figure \ref{fig:faces_overview}. Additionally, we were able to disclose a number of sensitive attributes (such as age or gender) associated with the sensitive data even in cases where the reconstructions were incomplete. We summarise our contributions as follows:

\begin{itemize}
    \item We explore the idea of using attacker-controlled data in a HbC setting to facilitate more accurate gradient-based reconstruction attacks;
    \item We evaluate a number of settings and models, showing quantitatively and qualitatively that our method outperforms attacks that rely on shared gradients alone;
    \item We present a number of augmentations to the existing model inversion algorithm that result in more accurate inverted images that can be used for further privacy-infringing downstream tasks, such as facial recognition or attribute inference attacks.
\end{itemize}

\section{Background and prior work}
Our work extends gradient-based model inversion attacks to allow the adversary to achieve better reconstruction performance. This type of attack was first described in Zhu et al. \cite{zhu2020deep} and allowed the adversary to reconstruct the training images in collaborative settings from shared gradient updates. However, this implementation was limited to images of low dimensions such as MNIST or CIFAR-10 and very shallow models, as otherwise the adversarial optimiser (L-BFGS) fails to converge, as it relies on the second-order derivatives, which are non-informative for functions such as ReLU (as the second-order derivative is constant), resulting in no reconstruction being produced. Additionally, in this implementation, both the label and the image have to be generated, forcing the adversary to perform two reconstruction tasks, significantly increasing the probability of an incorrect item being reconstructed. To alleviate this issue, Zhao et al \cite{zhao2020idlg} proposed \textit{improved deep leakage from gradients} (iDLG) that exploits the properties of the cross-entropy loss function allowing the adversary to always obtain the correct ground-truth label for the captured gradient. This is achieved through observation of the signs of the shared gradients with respect to the target label, as this label's corresponding gradient vector is negative, whereas every other label has a positive corresponding gradient vector. This method significantly improved the attack's performance, but does not scale to more complex models or datasets, with the L-BFGS optimiser still not converging in such settings for the aforementioned reasons. An advanced implementation of this attack was presented by Geiping et al., which allowed the attacker to reconstruct images with a size of $224\times224$ on deep models such as ResNets. The authors achieved this by proposing changes to both the adversarial optimiser, as well as to the gradient reconstruction cost function (replacing mean squared error with a cosine similarity loss). This allowed the adversary to target a larger number of learning contexts, yet certain limitations of the original method remained. The attack produces accurate results when the victim sends a single update per image, as otherwise the attacker is forced to attempt to reconstruct a batch of images corresponding to single gradient, which in most cases is an infeasible task even with this implementation. Work by Yin et al. \cite{yin2021see} allows to partially mitigate this issue through extraction of additional information about the training data from batch normalisation layers and allows reconstruction of higher quality on batches of up to $32$ images. This is done in a similar manner to \cite{zhao2020idlg}, by utilisation of signs of elements in a shared gradient vector, but this method is extended to a per-batch fashion, as authors observe that despite the aggregation of multiple samples, the large magnitude of the negative vectors allows them to remain identifiable even after the aggregation step. Thus, authors of \cite{yin2021see} propose to utilise the negative column-wise values in order to perform the label restoration procedure. However, this method makes an assumption that batch normalisation layers are present in the target model. While typically such layers are used in centralised ML, their application in the context of collaborative learning poses numerous challenges and requires the presence of such layers. Prior work has shown \cite{li2021fedbn} that BN is a poor choice in FL, thus this assumption may not be universally applicable. After reviewing other prior works in the area, we see a number of existing attack implementations that have an identical HbC adversary, but none of which investigate the priors that are available to such attacker. It is of note that concurrent work was recently published by \cite{hatamizadeh2022gradient}, which also exploits the adversarial priors similarly to our method, but is not conditioned on the activations or the style of the adversarial priors. 

Additionally, there exist other variations of model inversion attacks which exploit intermediate activations produced by the victim such as He et al. \cite{he2019model} where the central server is used in an inference mode and the adversary inverts the target activations to obtain the original image. This approach relies on a much stronger assumption of a compromised central server as well as relies on a single image being passed for inference. Additionally, an attack was proposed by Zhang et al. \cite{zhang2020secret} which relies on adversary having access to a suitable prior and a generative model in order to reconstruct the original image from the predictions returned by the target model. Similarly to He et al. \cite{he2019model}, the adversary is assumed to be a corrupted central server in an inference setting. While both of these attacks rely on a looser threat model, unlike the gradient-based reconstruction methods, they actually leverage adversarial priors and heavily depend upon the data that is available to the attacker.

\section{Methods}

\subsection{Threat model}
The general threat model of our work corresponds to the one in Geiping et al. \cite{geiping2020inverting} (i.e. IID (independent, identically distributed) FL setting with an adversarial client/central server that is able to capture a gradient update generated by their victim and has access to prior images from the same distribution as the victim data). We note that the adversary does \textit{not} need to know the ground-truth label of the image they are attacking in advance, as it can be extracted from the gradient as described in \cite{zhao2020idlg}.
\subsection{Gradient-based reconstruction}
The overview of the original gradient-only attack is identical to prior work by Geiping et al. \cite{geiping2020inverting} and can be summarised as follows:
\begin{enumerate}
    \item The adversary randomly generates an image-model update pair;
    \item The adversary captures the gradient update submitted by the victim;
    \item Using a suitable cost function (in our case cosine similarity), the adversary minimises the difference between the captured and the generated updates;
    \item The algorithm is repeated until the final iteration is reached.
\end{enumerate}
In \cite{geiping2020inverting}, the adversary attempts to reconstruct the image by solving the following minimisation problem:
\begin{equation*}
    \arg \min_{x'\in[0,1]^n} \left \{ 1 - \frac{\left \langle \nabla_{\theta}\mathcal{L}(x, y), \nabla_{\theta}\mathcal{L}(x', y)\right \rangle}{\Vert \nabla_{\theta}\mathcal{L}(x, y) \Vert_2 \cdot \Vert \nabla_{\theta}\mathcal{L}(x', y) \Vert_2} + \alpha \text{TV}(x)   \right\},
\end{equation*}
where $x'$ is the reconstruction target, $x$ is the ground truth, $y$ is the label, $\nabla_{\theta}\mathcal{L}$ is the gradient with respect to the weights, $\langle \cdot \rangle$ is the inner product in $\mathbb{R}^n$ and \mbox{$\Vert \cdot \Vert_2$} is the $L_2$-norm. $\alpha$ is a hyperparameter scaling the total variation penalty over the image, $\text{TV}(x)$ \cite{rudin1992nonlinear}. We set $\alpha = 10^{-6}$ in all experiments. We will refer to the aforementioned objective as the \textit{gradient loss} $l_g$.
\subsection{Activation matching}
The core idea of this work is in exploitation of additional information that is already available to the adversary. The core intuition behind our technique is that activations that belong to images of the same class are similar and conditioning the reconstruction on them can thus be used to guide the attack towards a result that looks similar to the original image.
\subsubsection{Activation matching}
A principled approach to penalise the images that look dissimilar from other images within the same class is to introduce an additional activation penalty term 
\begin{equation}
    \Vert A' - A \Vert_2^2 = {1\over N}\Vert \sum^{j}_{i=1} (a_i(x') - a_i(x))\Vert_2^2,
\end{equation}
where $A$ are the activations that correspond to the prior controlled by the attacker, $A'$ are the activations that correspond to the generated image and $a_j$ are activations at layer $j$ for the generated image ($x'$) and the adversarial prior ($x$) respectively. $N$ is the number of elements in $A$. In our case, we found that relying on an absolute (rather than squared) difference loss (the $L_1$ loss) results in better reconstructions, as outliers do not have a profound affect on the loss term. Therefore, the final \textit{activation loss} term is
\begin{equation}
    l_a \coloneqq \Vert A' - A \Vert_1 = {1\over N}\Vert \sum^{j}_{i=1} (a_i(x') - a_i(x))\Vert_1.
\end{equation}

We then combine the $l_g$ and $l_a$ into a single loss term that is then passed to the adversarial optimiser. Note that initially, no scaling is performed on either of the terms. We discuss the limitations of this approach in Section \ref{sec:results}, here we briefly note that while such approach still improves reconstruction across all settings, it does not result in significant improvements for those images that were not previously reconstructed using the baseline implementation. As a result, we turn our attention to investigating how these additional activation values can be best used by the adversary. 

\subsubsection{Scaling $l_a$}
\begin{figure}[h]
\centering

    \begin{subfigure}
    \centering
    \includegraphics[width=0.8\columnwidth]{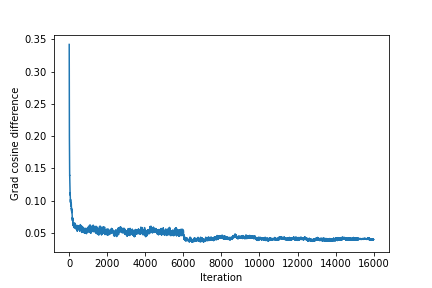}
    \end{subfigure}

    \begin{subfigure}
    \centering
    \includegraphics[width=0.8\columnwidth]{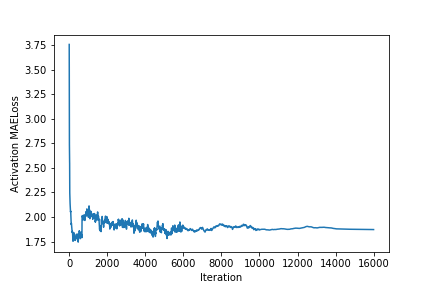}
    \end{subfigure}

\caption{Comparison of loss term magnitudes for gradient (above) and activation terms (below) (ResNet9, PPPD).}
\label{fig:graph_scaling_pppd}
\end{figure}

\begin{figure}[h]
\centering

    \begin{subfigure}
    \centering
    \includegraphics[width=0.8\columnwidth]{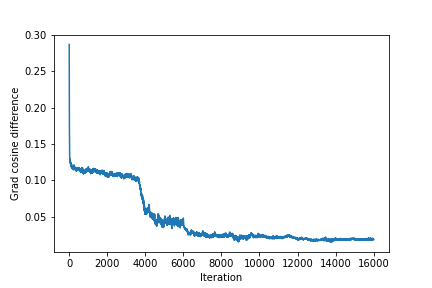}
    \end{subfigure}

    \begin{subfigure}
    \centering
    \includegraphics[width=0.8\columnwidth]{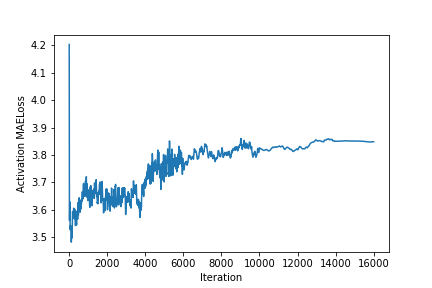}
    \end{subfigure}

\caption{Comparison of loss term magnitudes for gradient (above) and activation terms (below) (ResNet9, ImageNet).}
\label{fig:graph_scaling_imagenet}
\end{figure}

We notice that when comparing the penalty terms there is a clear imbalance with $l_a$ being approximately $10\times$ larger than the $l_g$ for the \textit{ConvNet} architecture (we use the \textit{ConvNet} name to describe the architecture described in \cite{geiping2020inverting}) and for ResNet architectures. As a result, we decided to experiment with various scaling factors in order to investigate the relationship between these two terms when it comes to assessing the reconstruction quality. Initially, we perform an analysis of the relative magnitudes of the individual terms in order to determine the suitable scaling factors. We report the results for ResNet9 on the ImageNet and Paediatric Pneumonia Prediction datasets (PPPD) (from \cite{kaissis2021end}) in Figures \ref{fig:graph_scaling_pppd} and \ref{fig:graph_scaling_imagenet}. We chose these datasets for exemplary purposes (representing single- and multi-channeled data), and similar results were obtained on other datasets. 

As evident from the magnitudes of the activation term weighted against the gradient penalty term, scaling is required to prevent one of the terms overpowering the contributions of the other. Thus, we selected a number of scaling factors $s_a$ and $s_g$ for both penalty terms where $s_a \in [0.1, 0.5, 1.0]$ and $s_g \in [1.0, 5.0, 10.0]$. We report results from these experiments in Section \ref{sec:results}. It is important to note that while this is an interesting addition to our main method, we do not perform any grid-search-based scaling constant selection and these were tuned manually. 

\subsection{Style reconstruction penalty term}
Additionally, when considering the problem of image reconstruction, we hypothesise that in certain contexts (such as a number of medical image analysis settings \cite{xu2019medical, burns2020machine, rajpurkar2017chexnet, elaziz2020new}) images of the same class only display a small amount of variation across different samples (i.e. minimal intra-class variation). Thus if an adversary possesses data that has a number of features very similar to the training sample, they would be able to extract it with higher fidelity as they have a strong prior. We perform this experiment similarly to the previous ones by creating a separate penalty term for the style loss that corresponds to style loss in a transfer learning setting \cite{johnson2016perceptual}. 

We fist define the \textit{Gram matrix} $G^{a}_{j}(x)$ as a $C_j \times C_j$ matrix whose elements are given by 
\begin{equation}
    G^{a}_{j}(x)_{c, c'} = \frac{1}{C_j \times H_j \times W_j} \sum^{H}_{h=1} \sum^{W}_{w=1} a(x)_{h, w, c} a(x)_{h, w, c'},
\end{equation}
where $a_{j}(x)$ are the activations of image $x$ at layer $j$, which is a feature map of shape $C_j \times H_j \times W_j$. The style loss term $l_s$ is then the squared Frobenius norm of the difference between the generated image and the adversarial image that produced the activations

\begin{equation}
    l_s \coloneqq \Vert G^{a}_{j}(x') - G^{a}_{j}(x)\Vert^{2}_{F},
\end{equation}
where $x$ is the adversarial prior and $x'$ is the generated image. 
\subsubsection{Scaling the combined terms}
\begin{figure}[h]
\centering

    \begin{subfigure}
    \centering
    \includegraphics[width=0.8\columnwidth]{figures/pppd_graph_grad.png}
    \end{subfigure}
    
    \begin{subfigure}
    \centering
    \includegraphics[width=0.8\columnwidth]{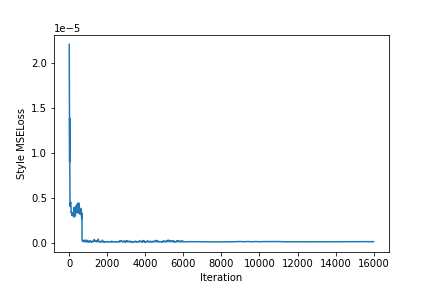}
    \end{subfigure}
    
\caption{Comparison of loss term magnitudes for gradient (above) and style terms (below) (ResNet9, PPPD).}
\label{fig:graph_style_pppd}
\end{figure}

\begin{figure}[h]
\centering

    \begin{subfigure}
    \centering
    \includegraphics[width=0.8\columnwidth]{figures/imagenet_graph_grad.png}
    \end{subfigure}
    
    \begin{subfigure}
    \centering
    \includegraphics[width=0.8\columnwidth]{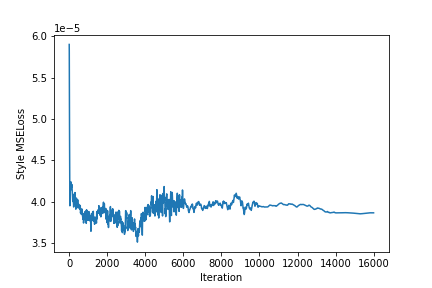}
    \end{subfigure}
    
\caption{Comparison of loss term magnitudes for gradient (above) and style terms (below) (ResNet9, ImageNet).}
\label{fig:graph_style_imagenet}
\end{figure}

Similarly to activation penalty term, we observe that style reconstruction loss is significantly lower than the gradient reconstruction loss (as seen in Figures \ref{fig:graph_style_pppd} and \ref{fig:graph_style_imagenet}) and we therefore perform scaling on this value in a similar manner. We note that on average (for CIFAR-10, ImageNet, FACES and BraTS) the gradient reconstruction term $l_g$ is $10^5\times$ larger than the style reconstruction term $l_s$ (for ConvNet- and ResNet models) and $10^6$ for VGG models. The gradient reconstruction term $l_g$ is also $10\times$ smaller than the activation loss term (with an exception of the ConvNet architecture, where it is $10^2\times$ smaller than the activation term). Thus, we select scaling factors $s_s$ for the style reconstruction term in a range of $[1.0, 10.0, 10\;000.0]$ and record our results for these values. One notable standalone architecture is ResNet34, for which no scaling is performed on the style penalty term and the activation penalty term is donwscaled by a factor of $10^2$ (irrespective of the main experimental setup). This is due to the fact that the magnitudes of the additional terms are significantly larger than the one of the gradient reconstruction term.

\subsubsection{Final reconstruction term}
The final (configurable) reconstruction loss term ($\mathcal{L}_r$) given an adversarial prior $x$ and the generated (or initialised) image $x'$ is given by: 
\begin{equation}
\mathcal{L}_r = s_g \cdot l_g + s_a \cdot l_a + s_s \cdot l_s,
\end{equation}
with $l_g$, $l_a$ and $l_s$ defined as above and $s_g$, $s_a$ and $s_s$ representing the corresponding scaling parameters.
With this loss term instead of the original gradient-based loss in \cite{geiping2020inverting}, we proceed to execute the reconstruction for $n$ iterations, where $n$ depends on the dataset but is generally in a range $[2\;000, 16\;000]$ iterations.

The overall quality of the reconstructed image is measured in line with prior work in terms of mean square error (MSE), peak signal-to-noise (PSNR) and structural similarity metric (SSIM) between the target image and the generated image similarly to prior work.

\begin{figure}[h!]
\centering
      \includegraphics[width=0.4\linewidth]{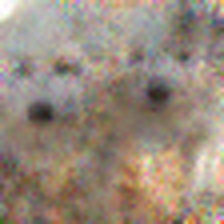}
    \hspace{5mm}%
      \includegraphics[width=0.4\linewidth]{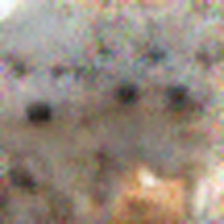}
\caption{Comparison of resulting reconstructions on ImageNet: ours (left) and baseline (right). Note that the baseline reconstruction is hallucinating $3$ dogs instead of $2$ present in the original image.}
\label{fig:dogs_comparison}
\end{figure}

\begin{figure}[h!]
\centering
      \includegraphics[width=0.4\linewidth]{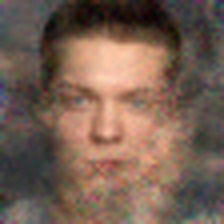}
    \hspace{5mm}%
      \includegraphics[width=0.4\linewidth]{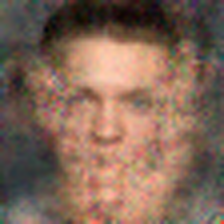}
\caption{Examples of reconstructions of the FACES dataset: our implementation (left) shows substantially more detail in the subject's mouth and facial periphery areas compared to the baseline (right).}
\label{fig:faces_comparison_more}
\end{figure}

\section{Experiments}
\label{sec:results}
In this work, we compare our method against the baseline gradient model inversion implementation by Geiping et al. \cite{geiping2020inverting}. Alternative implementations of the model inversion attack exist such as the original DLG and iDLG methods, but these were superseded by \cite{geiping2020inverting}. We also note that \cite{yin2021see, hatamizadeh2022gradient} have no source code publicly available and we were thus unable to reproduce their experiments. 

Our experimental setting was as follows: model architecture and datasets are selected in advance and shared in an IID manner across all participants. The batch size was set to $1$ unless specified otherwise. The optimiser used by both the federation and the adversary was AdamW \cite{loshchilov2019decoupled}, the learning rate for the federation and the adversary was set to $0.1$. Each adversarial class has between $3$ and $10$ images (the number determined randomly). The best reconstructions for our method were selected based on the lowest combined loss value for a given adversarial prior.

All experiments were performed on a system running Linux $20.04$ with $8$ CPU cores at $2.4$ GHz, $32$GB of RAM and an NVIDIA Quadro RTX $5000$ GPU.

We randomly selected a set of input images from the dataset not used for training (in batches of $32$) and allocated these as the adversarial prior. Based on the number of classes in the target dataset, the adversary had between $3$ and $12$ priors per each individual captured gradient. We ran our experiments for each image in a batch and report the average results as well as the average difference with reconstructions made that rely solely on the gradients. 

\subsection{Dataset descriptions}
In this work, we performed an attack on $2$ datasets identical to the ones deployed in \cite{geiping2020inverting,zhao2020idlg,yin2021see}, namely ImageNet ($224\times224$ images) and CIFAR-10 ($32\times32$ images). We additionally attack $3$ more learning settings, namely: (A) the PPPD dataset (images resized to $224\times224$, $3$ classes of normal/viral/bacterial), (B) the Brain Tumor Segmentation 2020 (BraTS) dataset \cite{menze2014multimodal} (images resized to $64\times64$ and the task changed to a binary classification of tumor/no tumor) and (C) the FACES \cite{ebner2010faces} dataset (images resized to $64\times64$, $6$ classes representing different emotions). For each of the \say{non-standard} tasks, we assign adversarial priors randomly: thus an adversary can have a T\textsubscript{1} or a T\textsubscript{2} image for BraTS for instance, as well as both a male and a female sample for FACES reconstruction. For smaller datasets (namely CIFAR-10, FACES and BraTS) the reconstruction was run for $2\;000$ iterations and for larger datasets (namely ImageNet and PPPD) the number of iterations was $16\;000$.

\subsection{Activation matching}
Here we present results of the attack when we use both the gradient loss term and the activation loss term in the generation procedure. We notice that even reliance on a simple approach that involves using the difference in activations as an unscaled and unnormalised additional penalty term, we managed to obtain more accurate reconstructions in a number of settings. Here, we experiment with a number of scaling factors in order to determine the relative importance of these on the quality of the reconstruction. We present an extensive overview of our results in Tables  (\ref{tab:activations_mean}) below. We note that across all experiments, our numerical results are largely contributing towards better reconstruction, while requiring a very minor adaptation of the reconstruction algorithm. 

\begin{table*}[t]
\centering
\caption{Reconstruction with an added activation matching term (32 images, CIFAR-10, ConvNet). Coefficients represent ($s_a, s_g, s_s$) respectively. We demonstrate the difference in various metrics when comparing images produced by our method against the ones produced by Geiping et al.'s method. An increase in the value is demonstrated by $\uparrow$ and a decrease in a value is demonstrated by $\downarrow$. Best values are highlighted like \emph{this.}}
\resizebox{\textwidth}{!}{%
\begin{tabular}{lccccccccc} 
\toprule
                       & \multicolumn{9}{c}{\textbf{Coefficients }}                                                                                                                                                \\
\multicolumn{1}{c}{}   & (1.0, 1.0, 0.0)   & (1.0, 5.0, 0.0)     & (1.0, 10.0, 0.0)    & (0.5, 1.0, 0.0)     & (0.5, 5.0, 0.0)     & (0.5, 10.0, 0.0)    & (0.1, 1.0, 0.0)   & (0.1, 5.0, 0.0)     & (0.1, 10.0, 0.0)     \\
MSE Difference (Mean)  & 0.0092 $\downarrow$ & 0.0188 $\downarrow$ & 0.0095 $\downarrow$ & 0.0125 $\downarrow$ & 0.0159 $\downarrow$ & 0.0151 $\downarrow$ & 0.0192$\downarrow$ & 0.0177  $\downarrow$ & \emph{0.0204} $\downarrow$  \\
PSNR Difference (Mean) & 0.3379 $\uparrow$  & 1.1389 $\uparrow$   & 0.3384 $\uparrow$   & 0.6117 $\uparrow$   & 0.6963 $\uparrow$   & 1.0023 $\uparrow$   & 1.2151 $\uparrow$  & 1.3249  $\uparrow$   & \emph{1.5508} $\uparrow$    \\
SSIM Differnce (Mean)  & 0.0110$\uparrow$   & 0.0412 $\uparrow$   & 0.0115 $\uparrow$   & 0.0192 $\uparrow$   & 0.0253 $\uparrow$   & 0.0264 $\uparrow$   & 0.0393 $\uparrow$  & 0.0411  $\uparrow$   & \emph{0.0579} $\uparrow$    \\
\bottomrule
\end{tabular}
\label{tab:activations_mean}
}
\end{table*}
\begin{table*}[t]
\centering
\resizebox{\textwidth}{!}{%
\begin{tabular}{lccccccccc} 
\toprule
                       & \multicolumn{9}{c}{\textbf{Coefficients }}                                                                                                                                                       \\
\multicolumn{1}{c}{}   & (1.0, 1.0, 0.0)   & (1.0, 5.0, 0.0)     & (1.0, 10.0, 0.0)    & (0.5, 1.0, 0.0)     & (0.5, 5.0, 0.0)     & (0.5, 10.0, 0.0)    & (0.1, 1.0, 0.0)   & (0.1, 5.0, 0.0)     & (0.1, 10.0, 0.0)     \\
MSE Difference (Max)  & 0.0633 $\downarrow$ & 0.1007 $\downarrow$ & 0.0633 $\downarrow$ & 0.0852 $\downarrow$ & 0.0910 $\downarrow$ & 0.0838 $\downarrow$ & 0.0740 $\downarrow$ & \emph{0.1093} $\downarrow$ & 0.0945 $\downarrow$  \\
PSNR Difference (Max) & 3.1867 $\uparrow$  & 5.4371 $\uparrow$   & 3.1867 $\uparrow$   & 3.4645 $\uparrow$   & 3.4049 $\uparrow$   & \emph{7.1717} $\uparrow$   & 3.2014 $\uparrow$  & 4.5041 $\uparrow$   & 3.8626 $\uparrow$    \\
SSIM Difference (Max)  & 0.1409 $\uparrow$   & 0.2130 $\uparrow$   & 0.1409 $\uparrow$   & 0.0946 $\uparrow$   & 0.1386 $\uparrow$   & 0.1785 $\uparrow$   & 0.1673 $\uparrow$  & \emph{0.2763} $\uparrow$   & 0.2022 $\uparrow$    \\
\bottomrule
\end{tabular}
}
\label{tab:activations_max}
\end{table*}

\begin{figure}[h!]
\centering
      \includegraphics[width=0.4\linewidth]{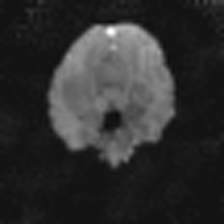}
    \hspace{5mm}%
      \includegraphics[width=0.4\linewidth]{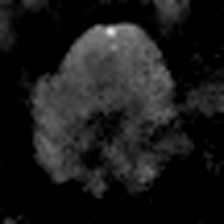}
\caption{Comparison of resulting reconstructions: ours (left) and baseline (right) (ResNet18, BraTS). While our method reconstructs the entire brain, the baseline method does not produce satisfactory results in the posterior brain regions and suffers from image artifacts around the brain.}
\label{fig:brats_1}
\end{figure}

\subsection{Activation matching with style penalty}
We observe that since the magnitude of the style transfer loss is negligible unless normalised and scaled, setting $s_s$ to $1.0$ only has a limited effect on the results of the reconstruction. As we discuss in Section \ref{sec:scale} however, experimenting with an increasing scaling factors allows us to further enhance the quality of the reconstructed image: both qualitatively and quantitatively in most cases. We report our findings in Table \ref{tab:style_mean}. As evident from results, appropriate scaling should be selected very carefully, taking the other two terms into account, as otherwise, the reconstruction quality can decrease. Additionally, we note, that under a biased non-IID distribution, there is a potential for the style loss term to alter the reconstructed image, which is undesirable. We discuss this further in Section \ref{sec:discussion}.
\begin{table*}[]
\centering
\caption{Scaling style penalty term (32 images, CIFAR-10, ConvNet). Coefficients represent ($s_a, s_g, s_s$) respectively.}
\resizebox{\textwidth}{!}{%
\begin{tabular}{lccccccc} 
\toprule
                       & \multicolumn{7}{c}{\textbf{Coefficients }}                                                                                                                                                       \\
\multicolumn{1}{c}{}   & (1.0, 1.0, 1.0)   & (1.0, 1.0, 100.0)     & (1.0, 1.0, 10000.0)    & (1.0, 10.0, 1.0)     & (1.0, 10.0, 100.0)     & (1.0, 10.0, 10000.0)         \\
MSE Difference (Mean)  & 0.0003 $\downarrow$ & 0.0021 $\downarrow$ & 0.0248 $\uparrow$ & 0.0258 $\downarrow$ & 0.0070 $\downarrow$ & \emph{0.0261} $\downarrow$ &   \\
PSNR Difference (Mean) & 0.4241 $\uparrow$  & 0.9973 $\uparrow$   & \emph{1.1906} $\downarrow$   & 0.8702 $\uparrow$   & 0.2295 $\uparrow$   & 0.8816 $\uparrow$   &     \\
SSIM Difference (Mean)  & 0.0349 $\uparrow$   & 0.0149 $\uparrow$   & 0.0254 $\downarrow$   & 0.0323 $\uparrow$   & 0.0126 $\uparrow$   &  \emph{0.0401} $\uparrow$       \\
\bottomrule
\end{tabular}
}
\label{tab:style_mean}
\end{table*}

\subsection{Scaling the penalty terms}
\label{sec:scale}
We adapted the same methodology as in the experiments above, while only altering the scaling factors by which the activations, the gradient and the style terms are multiplied to investigate the dependency between the separate penalty terms and the resulting image. We report the results in Table \ref{tab:scaling}. 

From the results above, we determine that scaling factors of ($1.0, 10.0, 10000.0$) for ($s_a, s_g, s_s$) result in the largest mean performance increase and we hence deploy these coefficients in the rest of the study.
\begin{table*}[]
\centering
\caption{Scaling individual penalty terms (32 images, CIFAR-10, ConvNet). Coefficients represent ($s_a, s_g, s_s$) respectively.}
\resizebox{\textwidth}{!}{%
\begin{tabular}{lccccccc} 
\toprule
                       & \multicolumn{7}{c}{\textbf{Coefficients }}                                                                                                                                                       \\
\multicolumn{1}{c}{}   & (1.0, 10.0, 5.0)   & (1.0, 10.0, 500.0)     & (1.0, 10.0, 50000.0)    & (1.0, 10.0, 1.0)     & (1.0, 10.0, 100.0)     & (1.0, 100.0, 10000.0)         \\
MSE Difference (Mean)  & 0.0004 $\downarrow$ & 0.0193 $\downarrow$ & 0.0017 $\uparrow$ & 0.0258 $\downarrow$ & 0.0070 $\downarrow$ & \textbf{0.0261} $\downarrow$ &   \\
PSNR Difference (Mean) & 0.1173 $\downarrow$  & 0.5547 $\uparrow$   & 0.6187 $\downarrow$   & 0.8702  $\uparrow$   & 0.2295 $\uparrow$   & \textbf{0.8816} $\uparrow$   &     \\
SSIM Difference (Mean)  & 0.0058 $\uparrow$   & 0.0277 $\uparrow$   & 0.0021 $\downarrow$   & 0.0323  $\uparrow$   & 0.0126 $\uparrow$   & \textbf{0.0401} $\uparrow$       \\
\bottomrule
\end{tabular}
}
\label{tab:scaling}
\end{table*}
\subsection{Targeting different architectures}
Finally, we perform a wider study to investigate how our method performs on a larger variety of network architectures. We specifically consider architectures such as VGG11, VGG13, VGG16, ResNet9, ResNet18 and ResNet34. We present results for CIFAR-10 in Table \ref{tab:arch_cifar}, for ImageNet in Table \ref{tab:arch_imagenet}, PPPD in Table \ref{tab:arch_pppd}, BraTS in Table \ref{tab:arch_brats}, and FACES in Table \ref{tab:arch_faces}. From these results we can deduce that our method outperforms the baseline in all settings, being particularly noticeable for ResNet-based architectures. In general, VGG-based architectures perform well in terms of relative quantitative difference with the baseline attack. However, for a large number of samples, neither the baseline nor our reconstructions represented meaningful inversions. We attribute this to the fact that the attack is hyper-sensitive to differences in initialisations for larger models. Attacks executed with a smaller number of iterations ($2\;000$ or less in comparison to $12\;000$ normally), but a larger number of restarts ($8$ instead of $1$) are significantly more likely to produce images that are better qualitatively as well as quantitatively. This principle is true for any architecture and dataset combination, but for VGG-based architectures a smaller number of restarts essentially rendered most reconstructions incorrect. This result is in line with prior work by \cite{ziller2021differentially} on segmentation models with VGG backbones.

\begin{table*}
\centering
\caption{Activation matching for various architectures (32 images, CIFAR10, $s_a=1.0, s_g=10.0, s_s=10000.0$)}
\begin{tabular}{lccccccc}
\toprule
                       & \multicolumn{6}{c}{\textbf{Architecture}}                                                                                        \\
\multicolumn{1}{c}{}   & ConvNet64         & VGG11               & VGG13               & VGG16      &ResNet9         & ResNet18            & ResNet34             \\
MSE Difference (Mean)  & 0.0095 $\downarrow$ & \emph{0.1247} $\downarrow$ & 0.0658 $\downarrow$ & 0.0528 $\downarrow$ & 0.0038 $\downarrow$ & 0.0159 $\downarrow$ & 0.0182 $\downarrow$  \\
PSNR Difference (Mean) & 0.3384 $\uparrow$  & 1.5311 $\uparrow$   & 0.2881 $\uparrow$   & 0.2150 $\uparrow$ & \emph{1.7756} $\uparrow$  & 0.6963 $\uparrow$   & 0.7261 $\uparrow$    \\
SSIM Differnce (Mean)  & 0.0115 $\uparrow$   & \emph{0.0726} $\uparrow$   & 0.0159 $\uparrow$   & 0.0043 $\uparrow$ & 0.0148 $\uparrow$  & 0.0253 $\uparrow$   & 0.0287 $\uparrow$    \\
\bottomrule
\end{tabular}
\label{tab:arch_cifar}
\end{table*}
\begin{table*}
\centering
\caption{Activation matching for various architectures (32 images, ImageNet, $s_a=1.0, s_g=10.0, s_s=10000.0$).}
\begin{tabular}{lccccccc} 
\toprule
                       & \multicolumn{6}{c}{\textbf{Architecture}}                                                                                        \\
\multicolumn{1}{c}{}   & ConvNet64         & VGG11               & VGG13               & VGG16      &ResNet9         & ResNet18            & ResNet34             \\
MSE Difference (Mean)  & 0.0088 $\downarrow$ & 0.0318 $\downarrow$ & 0.0329 $\downarrow$ & 0.0107 $\downarrow$ & \emph{0.2100} $\downarrow$ & 0.0078 $\downarrow$ & 0.0660 $\downarrow$  \\
PSNR Difference (Mean) & 0.1210 $\uparrow$  & 0.2514 $\uparrow$   & 0.1594 $\uparrow$   & 0.0257 $\uparrow$   & \emph{1.5366} $\uparrow$   & 0.2761 $\uparrow$ & 0.2753 $\uparrow$    \\
SSIM Differnce (Mean)  & 0.0081 $\uparrow$   & 0.0287 $\uparrow$   & 0.0170 $\uparrow$   & 0.0049 $\uparrow$   & \emph{0.0890} $\uparrow$   & 0.0183 $\uparrow$ & 0.0162 $\uparrow$    \\
\bottomrule
\end{tabular}
\label{tab:arch_imagenet}
\end{table*}
\begin{table*}
\centering
\caption{Activation matching for various architectures (32 images, PPPD, $s_a=1.0, s_g=10.0, s_s=10000.0$).}
\begin{tabular}{lccccccc} 
\toprule
                       & \multicolumn{6}{c}{\textbf{Architecture}}                                                                                        \\
\multicolumn{1}{c}{}   & ConvNet64         & VGG11               & VGG13               & VGG16    &ResNet9           & ResNet18            & ResNet34             \\
MSE Difference (Mean)  & 0.0227 $\downarrow$ & 0.0207 $\downarrow$ & \emph{0.0532} $\downarrow$ & 0.0529  $\downarrow$  & 0.0510  $\downarrow$ & 0.0316 $\downarrow$ & 0.0323 $\downarrow$  \\
PSNR Difference (Mean) & 0.6019 $\uparrow$  & 1.9772 $\uparrow$   & 0.3765 $\uparrow$   & 0.5785  $\uparrow$   & \emph{2.2235} $\uparrow$   & 1.1381  $\uparrow$ & 0.9124 $\uparrow$    \\
SSIM Differnce (Mean)  & 0.0445 $\uparrow$   & 0.0728 $\uparrow$   & 0.0332 $\uparrow$   & 0.0421 $\uparrow$   & \emph{0.0929} $\uparrow$   & 0.0398  $\uparrow$ & 0.0333 $\uparrow$    \\
\bottomrule
\end{tabular}
\label{tab:arch_pppd}
\end{table*}

\begin{figure}[h!]
\centering
      \includegraphics[width=0.4\linewidth]{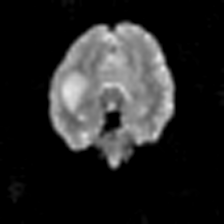}
    \hspace{5mm}%
      \includegraphics[width=0.4\linewidth]{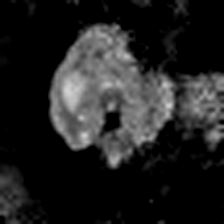}
\caption{Comparison of resulting reconstructions: ours (left) and baseline (right) (ResNet9, BraTS). Note the aliasing artifact on the left side of the brain (right side of the image) in the baseline method, which is missing a large part of the left frontal lobe (right side of the image).}
\label{fig:brats_2}
\end{figure}

\begin{table*}
\centering
\caption{Activation matching for various architectures (32 images, BraTS, $s_a=1.0, s_g=10.0, s_s=10000.0$).}
\begin{tabular}{lccccccc} 
\toprule
                       & \multicolumn{6}{c}{\textbf{Architecture}}                                                                                        \\
\multicolumn{1}{c}{}   & ConvNet64         & VGG11               & VGG13               & VGG16 &ResNet9               & ResNet18            & ResNet34             \\
MSE Difference (Mean)  & 0.0222 $\downarrow$ & 0.0850 $\downarrow$ & 0.0583 $\downarrow$ & 0.0001 $\downarrow$ & \emph{0.2560} $\downarrow$ & 0.0029 $\downarrow$ & 0.0163 $\downarrow$  \\
PSNR Difference (Mean) & 1.2576 $\uparrow$  & 1.6493 $\uparrow$   & 0.3282 $\uparrow$   & 0.2746 $\uparrow$   & \emph{5.4543} $\uparrow$ & 2.5245 $\uparrow$   & 0.5825 $\uparrow$    \\
SSIM Differnce (Mean)  & 0.0743 $\uparrow$   & 0.1048 $\uparrow$   & 0.0071 $\uparrow$   & 0.0142 $\uparrow$   & \emph{0.2428} $\uparrow$ & 0.0526 $\uparrow$   & 0.0521 $\uparrow$    \\
\bottomrule
\end{tabular}
\label{tab:arch_brats}
\end{table*}
\begin{table*}
\centering
\caption{Activation matching for various architectures (32 images, FACES, $s_a=1.0, s_g=10.0, s_s=10000.0$).}
\begin{tabular}{lccccccc} 
\toprule
                       & \multicolumn{6}{c}{\textbf{Architecture}}                                                                                        \\
\multicolumn{1}{c}{}   & ConvNet64         & VGG11               & VGG13               & VGG16 &ResNet9               & ResNet18            & ResNet34             \\
MSE Difference (Mean)  & 0.0241 $\downarrow$ & 0.0558 $\downarrow$ & 0.0317 $\downarrow$ & 0.0239 $\downarrow$ & 0.0933 $\downarrow$ & 0.0239 $\downarrow$ & \emph{0.1042} $\downarrow$  \\
PSNR Difference (Mean) & 0.4374 $\uparrow$  & 0.5039 $\uparrow$   & 0.1974 $\uparrow$   & 0.0614 $\uparrow$   & \emph{1.2821} $\uparrow$ & 0.2981 $\uparrow$   & 0.2997 $\uparrow$    \\
SSIM Differnce (Mean)  & 0.0409 $\uparrow$   & 0.0387 $\uparrow$   & 0.0108 $\uparrow$   & 0.0013 $\uparrow$   & \emph{0.0835} $\uparrow$ & 0.0022 $\uparrow$   & 0.0030 $\uparrow$    \\
\bottomrule
\end{tabular}
\label{tab:arch_faces}
\end{table*}

\subsection{Attacking at different stages of training}
\begin{figure}[h!]
\centering
      \includegraphics[width=0.4\linewidth]{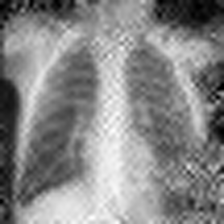}
    \hspace{5mm}%
      \includegraphics[width=0.4\linewidth]{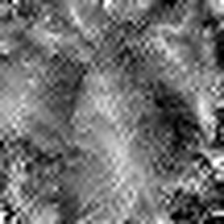}
\caption{Comparison of reconstructions on a model near convergence. Compared to the baseline reconstruction (right), which does not produce a satisfactory result, our method (left) achieves a clearly distinguishable image even when the model is trained.}
\label{fig:trained_figure}
\end{figure}

As discussed in \cite{geiping2020inverting}, gradient norms get significantly smaller, and thus, the gradients themselves are less descriptive towards the end of the training procedure. As a result, reconstructions are often unsuccessful for models that have previously been trained. We, therefore, investigate the effects our method has in these environments in order to determine if we are able to reconstruct images that were previously non-reconstructable by the adversary. Here we attack models trained on CIFAR-10 and PPPD datasets and compare the results to their untrained counterparts. We report the results in Table \ref{tab:trained} and exemplary reconstructions in Figure \ref{fig:trained_figure}.

\begin{table}
\centering
\caption{Attacks on previously trained models (32 images, ResNet-9, $s_a=1.0, s_g=10.0, s_s=10000.0$).}
\resizebox{0.5\textwidth}{!}{%
\begin{tabular}{lcc} 
\toprule
& CIFAR-10 (87.8\% Acc.) & PPPD (77.3\% Acc.)  \\
Ours (mean PSNR)     & 12.5 & 16.7 \\
Baseline (mean PSNR) & 11.2 & 15.7 \\
\bottomrule
\end{tabular}
}
\label{tab:trained}
\end{table}

In general, we see a trend similar to the results in \cite{geiping2020inverting}, showing that trained models are less susceptible to reconstruction attacks due to smaller and, thus, less informative gradient norms, resulting in significantly lower success ratio for the adversary. However, since we are able to leverage additional reconstruction terms that are independent from the gradient norm, we are able to produce reconstructions of higher quality even in such uninformative settings. In general, the number of successfully reconstructed images is greatly reduced, resulting in a number of noisy images that do not resemble the original training data.

\subsection{Downstream tasks based on the reconstructions}
While objective metrics of similarity between the original and the reconstructed image can be used effectively to asses the overall quality of reconstruction, it is sometimes the case that features of the image that are crucial for the main learning task do not have a significant contribution to the MSE between the two images. Therefore, in order to evaluate the effectiveness of certain tasks, we employed a number of downstream methods to assess the quality of our reconstructions for FACES datasets using alternative methods.

The first such task is an attribute inference attack, where we determine the approximate age and the attributes typically considered associated with the male or female gender of the victim through observation of the priors that were selected during the reconstruction. This essentially allows us to obtain sensitive characteristics of the victim regardless of the results of the reconstruction. The second downstream task is clustering of similar facial images based on \cite{schroff2015facenet}. For the purposes of this work, we explore the idea of utilisation of inverted images to spoof the results of the clustering task: We insert a number of images reconstructed by our method as well as the baseline method into a pool of data that contains facial images of the victim and perform clustering on this dataset. We then compare the $L_2$ distance between the embeddings produced from the images generated through our method and the baseline method. This distance can take values between $0$ (identical images) and $4.0$ (opposites of the spectrum), where the threshold of similarity lies between $1.0$ and $1.1$ (after which, in most cases the images can be considered dissimilar \cite{schroff2015facenet}). This allows us to use a comparison that relies on additional features rather than on the image similarity between the reconstruction and the original image, showing that our method is capable of producing data that is not just more accurate at reconstructions, but also produces images that can be utilised for additional adversarial tasks, such as facial spoofing \cite{ming2020survey} (i.e. tricking facial recognition systems). 

\subsubsection{Attribute inference from priors}
Out of $32$ randomly selected images for the ResNet-18 architecture, we are able to infer the approximate age in $22/32$ images. We are also able to infer the attributes typically considered associated with the male or female gender of the victim in $12/32$ cases. As we discuss below, while these results are very context specific, we nonetheless believe, that these findings show the additional threat associated with our model inversion method, allowing the adversary to infer the sensitive attributes of the training data regardless of the quality of the reconstruction itself. Additionally, we would like to point out that since most of the reconstructions were obtained from models that were untrained, certain features would not have the same impact on the activations as the others. As a result, we see a number of correctly predicted, but less privacy infringing characteristics (such as hair color) impacting other inference results (such as gender associated characteristics). Therefore, depending on the stage of the attack and the nature of the dataset, the effectiveness of this attack can be significantly improved in future work.

\subsubsection{Facial recognition clustering}
Finally, we perform facial clustering on the data generated by the two approaches as well as the original images in order to calculate the closeness between the reconstructions and the original data. For this we rely on the methodology from \cite{schroff2015facenet}, where we firstly identify if the images constitute valid faces (as reconstructions are selected at random and can significantly differ in quality), then produce a corresponding embedding and perform clustering on these accordingly. A detailed description of the procedure can be found in the original work by \cite{schroff2015facenet}. We provide an overview diagram of the attack in Figure \ref{fig:faces_overview}.

\begin{figure}[]
\centering
\includegraphics[width=0.5\linewidth]{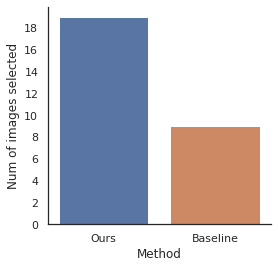}
\caption{Number of images selected as the closest neighbours from two sets of reconstructed images (higher is better).}
\label{fig:faces_selection}
\end{figure}

\begin{figure}[]
\centering
\includegraphics[width=0.5\linewidth]{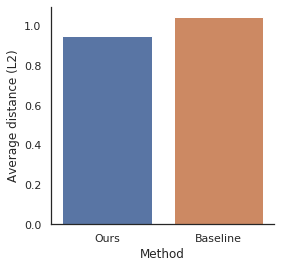}
\caption{Average $L_2$ distance between embeddings of the original images and the reconstructions (lower is better).}
\label{fig:faces_distance}
\end{figure}

We report our results in Figures \ref{fig:faces_selection} and \ref{fig:faces_distance}. It is of note that out of $32$ images only $28$ constituted valid faces (for both methods). As evident from these results, our method produces a significantly larger number of images that are similar to the original images ($19/28$ reconstructions). Furthermore, the $L_2$ distance between the original image embeddings and the ones generated by our method were, on average, $7.42\%$ lower when compared to the baseline images.

\section{Discussion}
\label{sec:discussion}
In this work we propose a novel formulation of gradient-based model inversion attack, where an adversary that is part of the training consortium uses their own data to guide the inversion process through matching the activations and the style of the victim's data. We see that this procedure allows the attacker to obtain reconstructions that are are quantitatively and qualitatively better than the baseline data-agnostic approach in all settings. During our experiments we have discovered a number of novel insights in the context of this attack.

\subsection{Not all models are created equal}
As we see from Section \ref{sec:results}, even within the same dataset, the results of the reconstruction are significantly affected by the selection of the shared model. There is a particularly noticeable divide between VGG-based architectures when compared to others: in general VGG models tend to produce much worse reconstructions and take significantly longer to produce any result that resembles the training data as seen in Figure \ref{fig:psnr_high_bad}. In our experimental setting, we hypothesise this comparative result arises due to $2$ factors. Firstly, ConvNet models (and in fact models that are more shallow than ConvNet, such as LeNet5 or an adaptation of LeNet5 by \cite{zhu2020deep} used in prior literature) are trivially invertible, as the number of parameters corresponding to a single image is much smaller ($2\;904\;970$ parameters) when compared to deeper models, resulting in fast adversarial convergence. Therefore when compared to such architectures, VGG (the smallest of which has $128\;807\;306$ parameters) nets perform worse due to their larger number of parameters and, thus, higher computational burden placed on the adversary. Secondly, while larger ResNet architectures are also significantly more complex (in terms of the width, depth and the overall number of trainable parameters) than their ConvNet counterparts, we hypothesise that these do not suffer from the same issues as the larger VGG-nets due to the better signal flow through the residual components, resulting in more \say{useful} information available to the adversary. Therefore, in general, the larger the model, the less likely the reconstruction to be successful, but certain architectural components, such as residual layers, have a potential to alleviate this issue to an extent. 
\subsection{Metrics are not everything}
\begin{figure}[]
\centering
\includegraphics[width=0.5\linewidth]{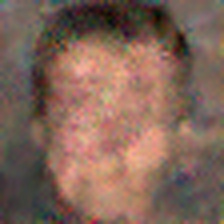}
\caption{Reconstruction (ours) with relatively high metrics (e.g. PSNR of $21.1346$), but missing facial features, rendering such reconstruction inaccurate.}
\label{fig:psnr_high_bad}
\end{figure}

While in a majority of cases, the reconstructions with higher PSNR and SSIM (when compared to the original image) can be considered accurate in representing the target image, there are certain cases, for which this conclusion does not always hold. This is of particular importance in datasets that put emphasis on smaller, but more significant features (e.g. facial features, rather than the information about the background around the face) for the main learning task. As such, situations such as the one described in Figure \ref{fig:psnr_high_bad} become possible, where both our method and the baseline method report PSNR values that are in line with other successful reconstructions in other datasets, but completely lacking important features that make full privacy violation possible. While we can attempt to identify gender characteristics and age from this image, any further privacy violation is simply impossible due to missing facial features, despite a relatively large PSNR and SSIM values when compared against the original image. This issue is exacerbated when comparing the results of VGG-nets against other models, as VGGs tend to have a higher relative MSE decrease, but are significantly harder to invert, resulting in better metrics, but poor reconstruction quality for both methods.

\subsection{Influence of the additional loss terms}
When discussing Figures \ref{fig:graph_scaling_imagenet} and \ref{fig:graph_style_imagenet} for instance, we can visually see that both additional loss terms experience a degree of fluctuation, where they are not minimised to the same level as the gradient difference loss term. This is because despite the fact that the adversarial image and the target image share a number of characteristics (by being part of the same class), there are some variations, thus, these two terms are expected to converge to a specific point, which is not required (and not even expected) to be a minima. Otherwise, we would expect the resulting reconstruction to resemble the adversarial image, rather than the target, as while these terms are scaled to the same magnitude, the style and the activation terms could dominate the reconstruction and dissolve the contribution of the gradient term. Therefore the behaviour we identify above for these terms is explainable. However, one potential adaptation that remains unexplored is the effect of scaling decay. In this work the scaling factors are pre-determined before the start of the training process and remain constant throughout. We hypothesise that their relative importance for the quality of the reconstruction might fluctuate, thus, making the exploration of individual term scaling decay an interesting investigation, potentially further enhancing the generated images further.
\subsection{Data complexity matters}
In this study we discuss two types of datasets: those with low and with high intra-class variations, where the former is represented in single-channel medical datasets and the latter in multi-channel datasets. We expect that the former dataset selection actually results in better reconstructions across the board, as the adversarial prior can often be similar to the target image both in features and style (due to the nature of medical data, as well as dependence only on a single channel) when compared to their multi-channel counterparts. While this conclusion is typically true qualitatively, this is not always the case quantitatively as seen from the comparison of certain extreme values in Tables \ref{tab:arch_cifar_max} and \ref{tab:arch_brats_max}. This happens for a number of factors that are not explicitly accounted for, such as the complexity of the task itself (as PPPD and BraTS are not only non-standard from a learning perspective in this study, but are also larger than CIFAR-10), larger penalty associated with the difference between two individual pixels etc. Therefore, while we can visually (and contextually) support the theory that settings with lower intra-class variation are much more vulnerable to this variation of model inversion attack, the relative difference between the generated and the original images in such setting can be a lot less significant when compared to datasets such as ImageNet or CIFAR-10. 

\subsection{Every stage of training is vulnerable}
One major limitation of the baseline approach is its dependence on the \say{descriptiveness} of the gradients i.e. reconstructions can lack in quality towards the end of the training procedure, as the norms of the associated gradients are much smaller. As a result, for larger datasets, such as ImageNet, the attack can fail much more often towards the end of the training procedure. However, as we show in Section \ref{sec:results}, it is still possible to attack the learning settings, where the model is already trained. While it is possible under certain circumstances to successfully reconstruct images in such settings with a baseline method (as reported by \cite{geiping2020inverting}), we note that our method achieves better results qualitatively and quantitatively, as well as being able to target models that are not accessible to the baseline approach. We attribute this to the fact that we do not rely on a single parameter (i.e. the gradient difference) alone, allowing the adversary to utilise $3$ reconstruction loss items, resulting in more informative reconstruction even in settings with limited gradient data available.

\subsection{On the selection of suitable priors}
In this work we did assigned adversarial datasets in a random fashion (only making sure that victim's data does not overlap with the adversarial data), thus limiting the advantage that they can have should the priors be selected with a specific criteria in mind. While in contexts such as pneumonia prediction, where images of the same class tend to be very similar to each other, other tasks such as ImageNet classification can feature images within the same class that do not share many features with each other. This is particularly important for BraTS learning task, as in our example, all modalities were available to the adversary and the best one was selected based on the lowest reconstruction loss value. However, we discovered that regardless of the results of the reconstruction (i.e. even in cases where the attack was unsuccessful for larger models) the correct adversarial prior always corresponded to the correct modality used by the victim. As a result, even if such attack never returns a complete reconstruction, the adversary essentially performs an attribute inference attack instead \cite{gong2016you}, where they can infer a sensitive attribute of the data record instead of reconstructing it in full. The impact on clients' privacy becomes significantly more profound for FACES dataset, since each class corresponds to an emotion, they all feature a participants of different genders and ages. We discovered that the adversary is able to infer both the age and the gender characteristics of victim's image irrespective of the results of the reconstruction, thus inferring attributes that could be sensitive otherwise. This task can be significantly more challenging in biased non-IID settings, where the adversary only has limited access to images of the same class or such images share very little with the target data (which could be the case for certain facial recognition tasks). While an in-depth investigation of \say{what makes a good prior good?} remains an open challenge, we still note that our method not only performs better than the baseline in most cases, but also results in an unintended disclosure of auxiliary attributes that belong to the victim even when the main attack fails and no meaningful reconstruction is returned. We consider an in-depth exploration of this property in different datasets with various architectures a promising area of future work.

\subsection{Future work}

While this variation of the model inversion attack achieves promising results, there exist a number of further adaptations that are currently outside of scope of this study, but are otherwise potentially beneficial for the adversary. The main problem we are facing when using additional penalty term is the issue of scaling: While in this work we already present quantitative and qualitative improvements over the baseline attack, we do not investigate the effects of various loss terms relative to each other. One promising approach that could help us to move away from arbitrary scaling to a more guided approach is multiple gradient descent algorithm (MGDA) used in adversarial backdoor synthesis. The ability to optimise for various learning tasks at once (i.e. to penalise the model separately for each individual loss term) would very likely lead to a significant improvement over our existing method. Furthermore, we note that certain layers in the model contribute to activations unequally (as discussed in \cite{zhang2019layers}: earlier layers can have a significantly more profound impact on the quality of the transferred, or in our case reconstructed, image), resulting in an additional optimisation challenge, which we will address in future work. In general, we believe that our method can benefit significantly from a more in-depth investigation into the domain of transfer learning in order to determine how to most effectively utilise the additional penalty terms and how to most efficiently perform the post-processing of the adversarial output to match the input image. Additionally, prior work outlines that the choice of a suitable metric for gradient comparison is a non-trivial task \cite{geiping2020inverting}, thus exploration of additional methods for this task remains an interesting challenge that we leave as future work. 

In this work, we concentrate explicitly on collaborative classification problems, but model inversion has previously been extended into other domains, such as medical image segmentation \cite{ziller2021differentially}. We believe that as the results of the prior work in the field are not yet consistent across all architectures and datasets (reconstructions often look distorted and incomplete) as well as the fact that our method performs well in homogeneous data distributions that can be associated with such studies, could make our method significantly improve the adversarial performance for such tasks. We leave an investigation of the domains beyond image classification (such as image segmentation) as future work.

\subsection{Limitations}

We consider the following limitations of our approach. Firstly, while the threat model remains identical to prior literature, we make an assumption that an adversary is placed in an IID setting, allowing them to control a small proportion of data that belongs to the same class as the victim's. Alternatively, we show that for certain datasets whose features do not experience a large inter-class variation, it is sufficient to only possess data that comes from the same distribution (e.g. pneumonia prediction task). As a result, when neither of these conditions are met, the attacker falls back to the original method and does not gain any benefit from this approach. For instance, we found that in a non-IID setting, utilisation of the style penalty term can reduce the accuracy of the reconstruction quantitatively without a consistent impact on the quality of the reconstruction. The second limitation is the computational requirements placed on the adversary, as in the current implementation of the selection algorithm each image of the same class as the victim's is used in a separate reconstruction procedure, requiring the adversary to run multiple reconstructions in order to obtain a single image of better quality. One mitigation of this limitation involves parallel computation of these reconstructions in order to select the best result. 

\section{Conclusion}
In this work, we propose a novel model inversion attack against collaborative machine learning, which shares the threat model with previously discussed gradient-based attacks, but offers more accurate reconstruction results. We achieve this by leveraging the data available to the adversary obtaining the activations associated with the data class that is shared with the other participants. We empirically demonstrate that even the untuned implementation of this algorithm yields better reconstruction results and has a significant potential for future application to various domains including transfer learning, and multi-objective optimisation. Additionally, we demonstrate the adverse effects our attack may have in real world contexts when applied to collaborative medical image segmentation or emotion prediction, where the datasets contain particularly sensitive information. We hope that this work can be used by both the privacy research community as well as the general machine learning community in order to better understand the adversarial perspective when designing collaborative systems and to facilitate the development of privacy-preserving machine learning systems.  

\bibliographystyle{plain}
\bibliography{references}
\newpage

\section{Appendix}
\begin{figure}[h!]
\centering
\includegraphics[width=0.8\linewidth]{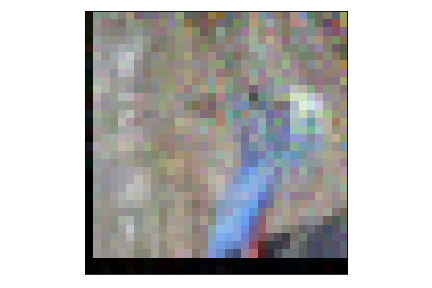}
\caption{Best reconstruction result: ours (above, PSNR of $42.1019$), baseline (below, PSNR of $21.3547$) (CIFAR-10, ResNet9, $s_a=1.0, s_g=10.0, s_s=1.0$, $6$ restarts, $2\;000$ iterations). Relative difference in MSE, PSNR and SSIM are $0.1156 \downarrow$, $20.7472 \uparrow$, $0.4381 \downarrow$ respectively.}
\label{fig:best}
\end{figure}

Here we present the best-case tables of our reconstruction against the baseline. As per Section \ref{sec:results}, an $\uparrow$ represents an increase in a value (e.g. PSNR) when compared to Geiping et al. \cite{geiping2020inverting} and a $\downarrow$ represents a decrease in a value. We additionally provide the absolute best-case reconstruction result with its corresponding parameters in Figure \ref{fig:best}.  

\begin{figure}[h!]
\centering
\includegraphics[width=0.8\linewidth]{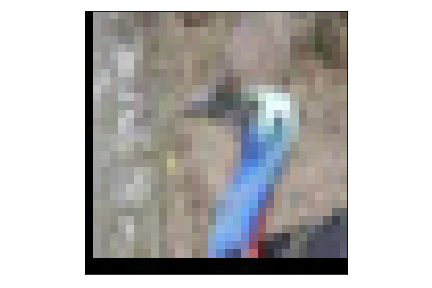}

\end{figure}

\begin{table*}
\centering
\caption{Activation matching for various architectures (32 images, CIFAR10, $s_a=1.0, s_g=10.0, s_s=10000.0$)}
\begin{tabular}{lccccccc}
\toprule
                       & \multicolumn{6}{c}{\textbf{Architecture}}                                                                                        \\
\multicolumn{1}{c}{}   & ConvNet64         & VGG11               & VGG13               & VGG16 & ResNet9              & ResNet18            & ResNet34             \\
MSE Difference (Mean)  & 0.0633 $\downarrow$ & \emph{0.7705} $\downarrow$ & 0.2833 $\downarrow$ & 0.2270 $\downarrow$ &
0.1156$\downarrow$ & 0.0159 $\downarrow$ & 0.0182 $\downarrow$  \\
PSNR Difference (Mean) & 3.1867 $\uparrow$  & 7.0077 $\uparrow$   & 1.2924 $\uparrow$   & 1.0595 $\uparrow$ & \emph{20.7472} $\uparrow$  & 0.6963 $\uparrow$   & 0.7261 $\uparrow$    \\
SSIM Differnce (Mean)  & 0.1409 $\uparrow$   & 0.4007 $\uparrow$   & 0.0844 $\uparrow$   & 0.0487 $\uparrow$ & \emph{0.4381} $\uparrow$  & 0.0253 $\uparrow$   & 0.0287 $\uparrow$    \\
\bottomrule
\end{tabular}
\label{tab:arch_cifar_max}
\end{table*}
\begin{table*}
\centering
\caption{Activation matching for various architectures (32 images, BraTS, $s_a=1.0, s_g=10.0, s_s=10000.0$).}
\begin{tabular}{lccccccc} 
\toprule
                       & \multicolumn{6}{c}{\textbf{Architecture}}                                                                                        \\
\multicolumn{1}{c}{}   & ConvNet64         & VGG11               & VGG13               & VGG16 &ResNet9               & ResNet18            & ResNet34             \\
MSE Difference (Max)  & 0.0341 $\downarrow$ & 0.1108 $\downarrow$ & 0.1053 $\downarrow$ & 0.0757 $\downarrow$ & \emph{0.4482} $\downarrow$ & 0.0031 $\downarrow$ & 0.0265 $\downarrow$  \\
PSNR Difference (Max) & 2.0597 $\uparrow$  & 2.9632 $\uparrow$   & 0.7108 $\uparrow$   & 1.2157 $\uparrow$   & \emph{9.3620} $\uparrow$ & 2.5959 $\uparrow$   & 0.7946 $\uparrow$    \\
SSIM Differnce (Max)  & 0.1176 $\uparrow$   & 0.1957 $\uparrow$   & 0.0171 $\uparrow$   & 0.0298 $\uparrow$   & \emph{0.4062} $\uparrow$ & 0.0598 $\uparrow$   & 0.0626 $\uparrow$    \\
\bottomrule
\end{tabular}

\label{tab:arch_brats_max}
\end{table*}
\begin{table*}
\centering
\caption{Activation matching for various architectures (32 images, ImageNet, $s_a=1.0, s_g=10.0, s_s=10000.0$).}
\begin{tabular}{lccccccc} 
\toprule
                       & \multicolumn{6}{c}{\textbf{Architecture}}                                                                                        \\
\multicolumn{1}{c}{}   & ConvNet64         & VGG11               & VGG13               & VGG16    &ResNet9           & ResNet18            & ResNet34             \\
MSE Difference (Max)  & 0.1168 $\downarrow$ & 0.2404 $\downarrow$ & 0.2357 $\downarrow$ & 0.1128 $\downarrow$ & \emph{1.2421} $\downarrow$ & 0.0941 $\downarrow$ & 0.1146 $\downarrow$  \\
PSNR Difference (Max) & 1.3219 $\uparrow$  & 1.3182 $\uparrow$   & 1.2513 $\uparrow$   & 0.4837 $\uparrow$   & \emph{4.8060} $\uparrow$ & 1.6867 $\uparrow$   & 1.3277 $\uparrow$    \\
SSIM Differnce (Max)  & 0.0760 $\uparrow$   & 0.1621 $\uparrow$   & 0.0815 $\uparrow$   & 0.0328 $\uparrow$   & \emph{0.2606} $\uparrow$ & 0.1240 $\uparrow$   & 0.0443 $\uparrow$    \\
\bottomrule
\end{tabular}
\label{tab:arch_imagenet_max}
\end{table*}
\begin{table*}
\centering
\caption{Activation matching for various architectures (32 images, PPPD, $s_a=1.0, s_g=10.0, s_s=10000.0$)}
\begin{tabular}{lccccccc} 
\toprule
                       & \multicolumn{6}{c}{\textbf{Architecture}}                                                                                        \\
\multicolumn{1}{c}{}   & ConvNet64         & VGG11               & VGG13               & VGG16   &ResNet9            & ResNet18            & ResNet34             \\
MSE Difference (Max)  & 0.1308 $\downarrow$ & 0.0576 $\downarrow$ & \emph{0.3085}  $\downarrow$ & 0.1168  $\downarrow$  & 0.1301  $\downarrow$ & 0.0586 $\downarrow$ & 0.0644 $\downarrow$  \\
PSNR Difference (Max) & 2.8425 $\uparrow$  & \emph{5.4895} $\uparrow$   & 2.1320 $\uparrow$   & 2.3886  $\uparrow$  & 4.6085  $\uparrow$  & 1.9259 $\uparrow$   & 3.2286 $\uparrow$    \\
SSIM Differnce (Max)  & 0.1891 $\uparrow$   & 0.2210 $\uparrow$   & 0.1964 $\uparrow$   & 0.1456 $\uparrow$  & \emph{0.2548}  $\uparrow$  & 0.1438 $\uparrow$   & 0.1158 $\uparrow$    \\
\bottomrule
\end{tabular}
\label{tab:arch_pppd_max}
\end{table*}
\begin{table*}
\centering
\caption{Activation matching for various architectures (32 images, FACES, $s_a=1.0, s_g=10.0, s_s=10000.0$).}
\begin{tabular}{lccccccc} 
\toprule
                       & \multicolumn{6}{c}{\textbf{Architecture}}                                                                                        \\
\multicolumn{1}{c}{}   & ConvNet64         & VGG11               & VGG13               & VGG16 &ResNet9               & ResNet18            & ResNet34             \\
MSE Difference (Max)  & 0.0799 $\downarrow$ & 0.1785 $\downarrow$ & 0.1238 $\downarrow$ & 0.0239 $\downarrow$ & 0.3472 $\downarrow$ & 0.1350 $\downarrow$ & \emph{0.5757} $\downarrow$  \\
PSNR Difference (Max) & 1.4266 $\uparrow$  & 1.8020 $\uparrow$   & 0.7910 $\uparrow$   & 0.0614 $\uparrow$   & \emph{4.4027} $\uparrow$ & 2.1986 $\uparrow$   & 1.8718 $\uparrow$    \\
SSIM Differnce (Max)  & 0.1029 $\uparrow$   & 0.1656 $\uparrow$   & 0.0383 $\uparrow$   & 0.0013 $\uparrow$   & \emph{0.3054} $\uparrow$ & 0.1118 $\uparrow$   & 0.0396 $\uparrow$    \\
\bottomrule
\end{tabular}
\label{tab:arch_faces_max}
\end{table*}

\end{document}